\pdfoutput=1
\documentclass[letterpaper, 10 pt, journal, twoside]{IEEEtran}

\IEEEoverridecommandlockouts
\usepackage{graphicx}
\usepackage{multirow}
\usepackage{booktabs}
\usepackage{amsmath}
\usepackage{amssymb}
\usepackage{colortbl}
\usepackage[x11names]{xcolor}
\usepackage{cite}
\usepackage{algorithm}
\usepackage{algpseudocode}
\usepackage[font=small]{caption} %
\usepackage{threeparttable}
\usepackage[numbers,sort&compress]{natbib}

\usepackage{threeparttable}  %
\usepackage{url}      %
\usepackage{hyperref}  %
\usepackage{float}
\usepackage{placeins} %
\hypersetup{
    colorlinks=true,
    linkcolor=blue,
    filecolor=magenta,
    urlcolor=blue,
    citecolor=blue
}

\title{\LARGE \bf
MS-Occ: Multi-Stage LiDAR-Camera Fusion for 3D Semantic Occupancy Prediction}

\author{Zhiqiang Wei, Lianqing Zheng, Jianan Liu$^{\dagger}$, Tao Huang, ~\IEEEmembership{Senior Member,~IEEE,}\\ Qing-Long Han, ~\IEEEmembership{Fellow,~IEEE,} Wenwen Zhang, ~\IEEEmembership{Member,~IEEE,} Fengdeng Zhang
\thanks{ $^{\dagger}$ Corresponding Author.}
\thanks{Zhiqiang Wei and Fengdeng Zhang are with the School of Optical-Electrical and Computer Engineering, University of Shanghai for Science and Technology, Shanghai 200093, China. Email: 142460402@st.usst.edu.cn, FDZhang@usst.edu.cn}
\thanks{Lianqing Zheng is with the School of Automotive Studies, Tongji University, Shanghai 201804, China. Email: zhenglianqing@tongji.edu.cn.}
\thanks{Jianan Liu is with Momoni AI, Gothenburg, Sweden. Email: jianan.liu@momoniai.org.}
\thanks{Tao Huang is with College of Science and Engineering, James Cook University, Cairns, QLD 4878, Australia. Email: tao.huang1@jcu.edu.au.}
\thanks{Qing-Long~Han is with the School of Engineering, Swinburne University of Technology, Melbourne, VIC 3122, Australia. Email: qhan@swin.edu.au.}
\thanks{Wenwen Zhang is with School of Electrical and Electronic Engineering, Nanyang Technological University, Singapore 639798, Singapore. Email: wenwen.zhang@ntu.edu.sg.}
}

\begin{document} 

\maketitle

\begin{abstract}
Accurate 3D semantic occupancy perception is essential for autonomous driving in complex environments with diverse and irregular objects. While vision-centric methods suffer from geometric inaccuracies, LiDAR-based approaches often lack rich semantic information. To address these limitations, MS-Occ, a novel multi-stage LiDAR-camera fusion framework which includes middle-stage fusion and late-stage fusion, is proposed, integrating LiDAR's geometric fidelity with camera-based semantic richness via hierarchical cross-modal fusion. The framework introduces innovations at two critical stages: (1) In the middle-stage feature fusion, the Gaussian-Geo module leverages Gaussian kernel rendering on sparse LiDAR depth maps to enhance 2D image features with dense geometric priors, and the Semantic-Aware module enriches LiDAR voxels with semantic context via deformable cross-attention; (2) In the late-stage voxel fusion, the Adaptive Fusion (AF) module dynamically balances voxel features across modalities, while the High Classification Confidence Voxel Fusion (HCCVF) module resolves semantic inconsistencies using self-attention-based refinement. 
Experiments on two large-scale benchmarks demonstrate state-of-the-art performance.
On nuScenes-OpenOccupancy, MS-Occ achieves an Intersection over Union (IoU) of $32.1\%$ and a mean IoU (mIoU) of $25.3\%$, surpassing the state-of-the-art by $+0.7\%$ IoU and $+2.4\%$ mIoU. 
Furthermore, on the SemanticKITTI benchmark, our method achieves a new state-of-the-art mIoU of $24.08\%$, robustly validating its generalization capabilities.
Ablation studies further confirm the effectiveness of each individual module, highlighting substantial improvements in the perception of small objects and reinforcing the practical value of MS-Occ for safety-critical autonomous driving scenarios.
\end{abstract}



\section{Introduction}
3D semantic occupancy is vital for autonomous driving, enabling a comprehensive understanding of complex environments for downstream tasks such as vehicle planning and control. Unlike traditional 3D object detection, occupancy perception captures irregular and non-rigid objects beyond predefined categories. Vision-based 3D semantic occupancy perception has shown notable performance across various technical paradigms, including monocular setups \cite{monoscene}, surround-view camera systems \cite{surroundocc, bevformer, FB-BEV}, and efficient architectural designs \cite{fullySparseOcc, FastOcc}. However, vision-based methods face inherent challenges like inaccurate depth estimation, limited geometric fidelity, and sensitivity to lighting conditions, undermining robustness in real-world driving environments. These challenges highlight the importance of integrating LiDAR and camera data to improve occupancy perception. Although 4D radar has been explored in recent studies \cite{OmniHDScenes, Doracamom} as a supplement to mitigate some of these issues, it remains constrained by limitations such as point cloud noise and sparsity.

\begin{figure}[t]
    \centering
    \includegraphics[width=0.8\linewidth]{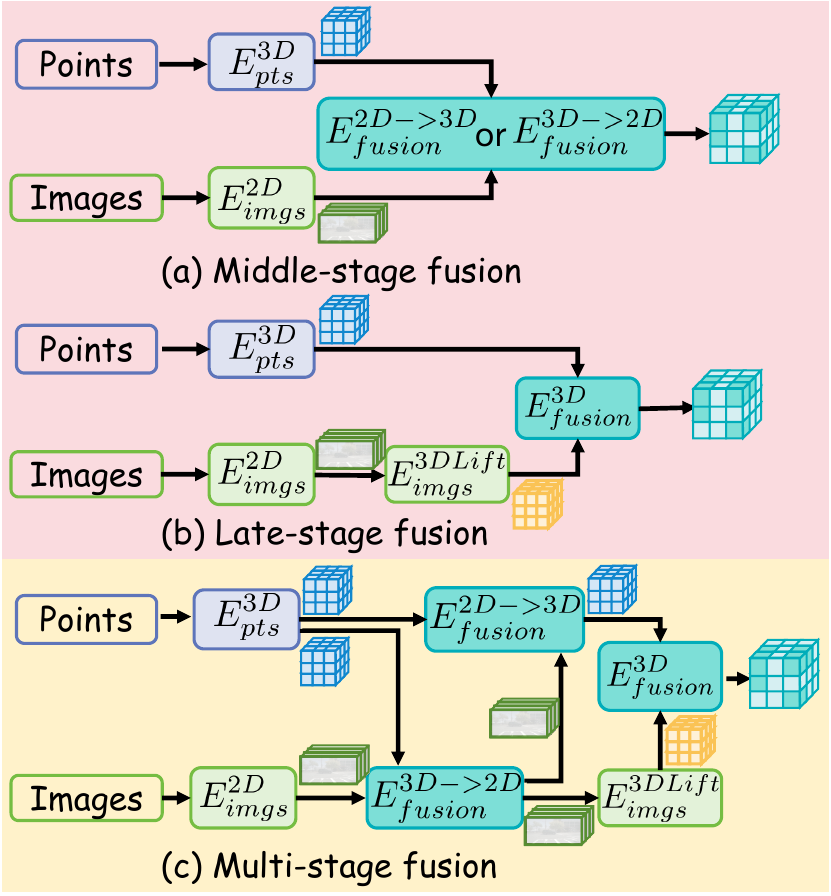}
    \caption{Comparison of occupancy fusion stages in encoder design. (a) Middle-stage fusion (e.g., \cite{occloff}), where 2D image features are directly fused with 3D LiDAR voxel features. (b) Late-stage fusion, including methods such as \cite{wang2023openoccupancy, Co-Occ, OccGen, EFFOcc}, which explicitly lift 2D image features to 3D voxel features (e.g., using \cite{lss}), followed by 3D fusion in voxel space. (c) Multi-stage fusion, as proposed in our MS-Occ, integrates both middle and late-stage fusion within a hybrid architecture.}
    \label{fig:fusion_method}
    \vspace{-12pt}
\end{figure}

LiDAR-camera integration is a key research focus in multi-modal sensor fusion. Multi-modal 3D semantic occupancy perception can be classified into four fusion stages based on the timing of feature integration \cite{MultiModalObjectDetection, multilayer_depthEstimationRC}: early-stage fusion, middle-stage fusion, late-stage fusion, and multi-stage fusion. Early-stage fusion involves the combination of raw or preprocessed sensor data, allowing for full utilization of the original information. However, it is prone to misalignment caused by modality-specific limitations and calibration errors. As a result, most existing methods prioritize middle or late-stage fusion, as illustrated in Fig.~\ref{fig:fusion_method}(a) and (b).

Middle-stage fusion integrates 2D image features with 3D LiDAR features through projection-based fusion $E_{\text{fusion}}^{\text{3D}\rightarrow\text{2D}}$ or cross-attention mechanisms $E_\text{fusion}^{\text{2D}\rightarrow\text{3D}}$. For instance, OccLoff \cite{occloff} employs cross-attention to fuse 2D image features with LiDAR voxel features. Nonetheless, middle-stage fusion methods often fail to fully exploit the strengths of each modality within the 3D voxel space. In contrast, late-stage fusion combines 3D image voxel features, explicitly lifted from 2D image features ($E_\text{imgs}^\text{3DLift}$), with 3D LiDAR voxel features in voxel space ($E_\text{fusion}^\text{3D}$). Representative examples include M-CONet \cite{wang2023openoccupancy} and Co-Occ \cite{Co-Occ}, both of which enhance performance through late-stage fusion. OpenOccupancy \cite{wang2023openoccupancy} presents a novel LiDAR-camera fusion baseline that leverages the complementary advantages of the sensors and establishes the first benchmark for fusion-based surrounding occupancy perception. Co-Occ \cite{Co-Occ} further refines LiDAR features by incorporating neighboring camera features using a KNN search strategy in GSFusion.

Despite their success, these approaches perform fusion solely in voxel space, overlooking the hierarchical nature of 2D image features. This omission introduces geometric and semantic biases during voxel grid construction, thereby limiting the alignment of multi-modal features in the 3D voxel domain. To address these limitations, multi-stage fusion, illustrated in Fig.~\ref{fig:fusion_method}(c), combines both middle- and late-stage fusion strategies. This method leverages both 2D visual and 3D LiDAR features for a more robust alignment of semantic richness and geometric precision. Nevertheless, multi-stage fusion remains underexplored in existing research.

This study introduces MS-Occ, to our knowledge, the first LiDAR-camera fusion framework for 3D semantic occupancy prediction using multi-stage fusion.
Initially, features are extracted from both LiDAR point clouds and surround-view images.
These features are then enhanced through middle-stage feature fusion, where 2D camera features are enriched with geometric information using the proposed Gaussian-Geo module, and LiDAR features are augmented with semantic cues via the Semantic-Aware module.
In the Gaussian-Geo module, a Gaussian kernel is applied to produce a denser LiDAR heatmap, which interacts with the original 2D image features to generate geometrically enhanced camera features.
These enhanced features are subsequently fused with 3D LiDAR features using deformable attention in the semantic-aware module, thereby injecting semantic context into the LiDAR representations.

Following this middle-stage fusion, a late-stage voxel fusion is performed to further refine the fused voxel features from both modalities.
This stage comprises two complementary components: the High Classification Confidence Voxel Fusion (HCCVF) module and the Adaptive Fusion (AF) module.
The AF module initially performs adaptive fusion of voxel features, following the fusion method of M-CONet \cite{wang2023openoccupancy}, to integrate voxel representations from both LiDAR and camera streams.
The innovative HCCVF Module then selectively refines high-confidence features from each modality through an attention mechanism, thereby mitigating semantic ambiguities between them.
The final fused voxel features are passed through a prediction head to generate the 3D semantic occupancy output.

The main contributions are summarized as follows:
\begin{itemize} \setlength\itemsep{-3pt}
\item 
A novel LiDAR-camera multi-stage fusion framework for 3D semantic occupancy prediction is proposed. To our knowledge, this is the first multi-stage fusion architecture for this task. The framework efficiently processes LiDAR and camera data to generate 3D voxel representations that compensate for modality-specific limitations.

\item 

Four dedicated modules are introduced for unified and fine-grained perception. The Gaussian-Geo module enriches camera features with dense geometric priors. The Semantic-Aware module transfers image semantics to LiDAR voxels through deformable attention. In the late-stage, the HCCVF and AF modules are integrated: HCCVF resolves semantic ambiguities via self-attention, while the AF module dynamically balances geometric and semantic voxel features.

\item 
Extensive experiments on the nuScenes-OpenOccupancy and SemanticKITTI benchmarks demonstrate that MS-Occ outperforms existing state-of-the-art methods, achieving superior geometric and semantic accuracy, with significant improvements in safety-critical categories.
\end{itemize}

\begin{figure*}[t]
    \centering
    \includegraphics[width=0.8\linewidth]{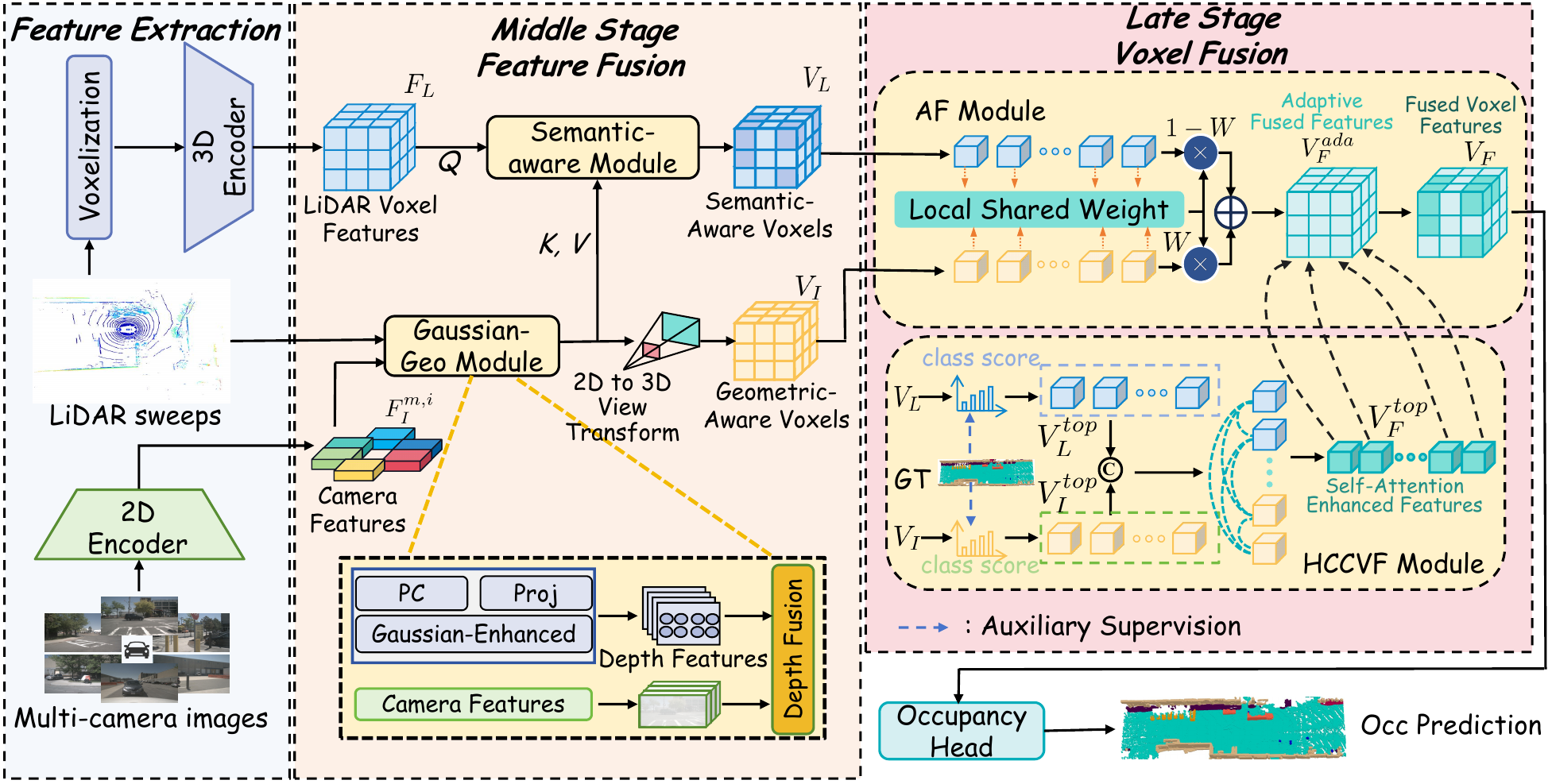}
    \caption{The overall framework of MS-Occ, with the four proposed modules highlighted in \textbf{light yellow}. The pipeline consists of two main stages: middle-stage fusion and late-stage fusion. In the middle-stage fusion, camera and LiDAR features are fused to produce geometry-enhanced image features via the Gaussian-Geo module, and semantically enriched LiDAR features via the Semantic-Aware module. Subsequently, in the late-stage fusion, the AF module and the HCCVF module are applied in parallel to integrate the fused representations and generate the final 3D semantic occupancy grid of the scene.}
    \label{fig:framework}
    \vspace{-10pt}
\end{figure*}

\section{Related Works}
\subsection{Vision-based 3D Occupancy Prediction}
Vision-based occupancy perception has seen significant recent advances. Transforming 2D image features into 3D representations (e.g., Bird’s-Eye View (BEV) features \cite{yu2023flashocc}, Tri-Perspective View (TPV) features \cite{TPVFormer}, and voxel features \cite{surroundocc}) is critical for this task. 
Three dominant approaches have emerged for this transformation \cite{occ_survey}: 
(1) The projection-based method \cite{monoscene, miao2023occdepth} employs perspective projection models to establish geometric correspondences by projecting voxel centroids from 3D space onto 2D image planes.
(2) The depth prediction method \cite{zhang2023occformer, yu2023flashocc, FastOcc}, derived from the Lift-Splat-Shoot (LSS) paradigm, uses estimated depth distributions to lift 2D features into 3D space. 
(3) The cross-attention method \cite{surroundocc, wang2024panoocc, TPVFormer} facilitates feature interaction between 3D voxel grids and 2D feature maps through deformable attention \cite{DeformableDETR}. 
Recent works such as SparseOcc \cite{fullySparseOcc} and EFFOcc \cite{EFFOcc} have also improved computational efficiency.
However, vision-based 3D occupancy perception faces fundamental limitations due to sensitivity to illumination changes, inherent depth estimation errors, and compromises in geometric precision.

\subsection{LiDAR-Camera Fusion for 3D Occupancy Prediction}
LiDAR-camera fusion has shown exceptional performance via four primary strategies \cite{occ_survey}: 
(1) Concatenation-based methods \cite{occfusion, Co-Occ}, which concatenate 3D feature volumes from distinct modalities.
(2) Summation-based approaches \cite{wang2023openoccupancy, OccGen}, which perform element-wise additive operations to combine geometrically aligned sensor representations.
(3) Attention-based methods \cite{occloff}, inspired by SparseFusion \cite{sparseFusion} and TransFusion \cite{TransFusion}, which apply cross-modality attention mechanisms. 
(4) Projection-based methods \cite{mseg3d_cvpr2023, Huang2020EPNet}, which align LiDAR points with image pixels via calibrated projection.
Recent research fuses RGB semantics with depth for dense scene reconstruction~\cite{rosinol2021kimera, ilabel, zheng2024semantic}, further highlighting the shift toward semantically enriched geometry.

However, these strategies primarily focus on single middle- or late-stage fusion mechanisms, neglecting fusion timing and thereby constraining dynamic alignment of representations across modalities.
To address these challenges, we introduce a two-stage enhancement and fusion strategy.
In the middle stage, camera features are enriched with geometric priors through lightweight Gaussian kernel rendering (Gaussian-Geo module),
which provides dense cues without requiring heavy separate depth networks,
while LiDAR features are augmented with semantic context using efficient deformable attention (Semantic-Aware module),
which adaptively queries salient information more efficiently than global attention.
In the late stage, voxel-level fusion is performed by first applying adaptively weighting (AF module) and subsequently refining the result through targeted self-attention (HCCVF module) refines high-confidence or conflicting voxels to resolve semantic inconsistencies, a more robust alternative than conventional summation or simple concatenation that treats all voxels equally.
The proposed MS-Occ leverages modality-specific features at multiple fusion stages to facilitate cross-modal alignment and enhance the quality of fused representations.

\section{Method}
The overall architecture is illustrated in Fig.~\ref{fig:framework}. Modality-specific features are acquired using 2D and 3D encoders. 
The encoded 2D image features, are denoted as $F_\mathrm{I}^{(m,i)} \in \mathbb{R}^{H_\mathrm{I} \times W_\mathrm{I} \times C_\mathrm{I}}$ , where $m$ and $i$ indicate the $m$-th camera view and $i$-th image scale, respectively. 
Here, $H_\mathrm{I}$ and $W_\mathrm{I}$ represent the spatial dimensions of the image feature map, and $C_\mathrm{I}$ denotes the feature channel dimension. 
The sparse 3D LiDAR voxel features are denoted as $F_{\mathrm{L}} \in \mathbb{R}^{D \times H \times W \times C}$, where $D$, $H$, $W$ are the depth, height, and width of the voxel grid, and $C$ is the number of feature channels for each voxel.

\subsection{Middle Stage Feature Fusion} \label{middl-fusion}
Although the proposed method aims to leverage the advantages of both modalities, it is important to address their respective limitations. Specifically, the LiDAR modality struggles to capture rich semantic information, while the camera modality lacks precise geometric and depth estimation capabilities. To mitigate these modality-specific limitations at the intermediate feature level, novel geometric and semantic information fusion modules are introduced and integrated into the network.

\subsubsection{Gaussian-Geo Module} \label{gaussiangeo-net}
\begin{figure}[t]
    \centering
    \includegraphics[width=0.8\linewidth]{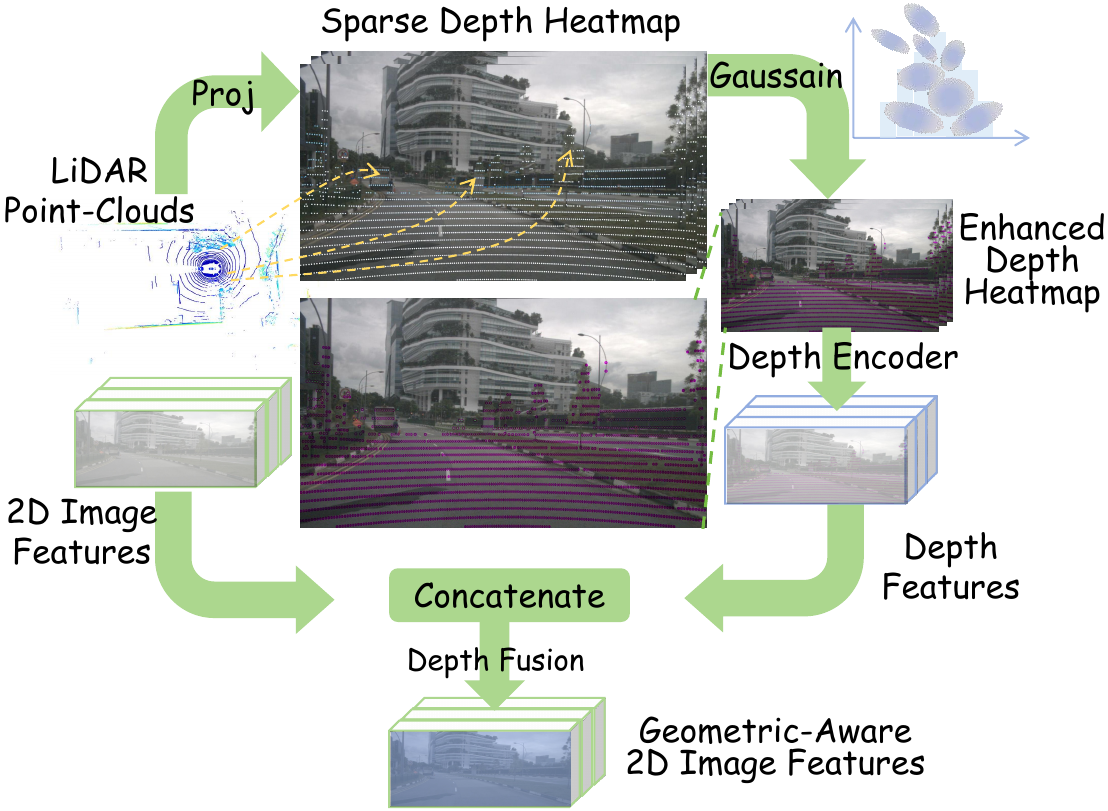}
    \caption{Illustration of intermediate outputs from the proposed Gaussian-Geo module. The LiDAR point cloud is densified using Gaussian kernel, and the resulting geometric information is subsequently transferred to the image to enhance its spatial representation.}
    \label{fig:gaussianGeo-Net}
    \vspace{-12pt}
\end{figure}

The Gaussian-Geo module is designed to incorporate geometric priors through Gaussian kernel rendering, drawing inspiration from recent approaches \cite{gaussianocc, gaussianFormer} that utilize 3D Gaussians \cite{3DGS} for scene representation, as illustrated in Fig.~\ref{fig:gaussianGeo-Net}.
Specifically, depth maps are first obtained from LiDAR point clouds. A straightforward approach involves directly projecting point clouds onto images using intrinsic and extrinsic calibration parameters to generate depth maps. However, due to the inherent sparsity of LiDAR data, the resulting depth maps are typically incomplete and lack detailed features. 
To mitigate this limitation, a Gaussian kernel rendering technique is introduced to enhance the projected depth maps across multi-view images, resulting in more informative depth representations $Y \in [0, U]^{H_\mathrm{C} \times W_\mathrm{C} \times 1}$, where $U$ denotes the LiDAR point depth value, $H_\mathrm{C}$ and $W_\mathrm{C}$ represent the height and width of the original camera image, respectively. 
The raw LiDAR points are splatted onto the image planes of multiple views. For each hit position $p_g \in \mathbb{R}^\mathrm{2}$ as a center, a Gaussian kernel with the corresponding depth value is generated in the image plane. The final depth value at position $q \in \mathbb{R}^\mathrm{2}$ is computed as:
\begin{equation}
    Y_q = \underset{g \in \mathcal{G}_\text{hit}}{\min} \left( U^g \mathtt{exp}\left( -\frac{\left( q - p_g \right)^2}{\mathrm{2\sigma^2}} \right) \right),
\end{equation}
where $\sigma$ controls the spread of the kernel, $\mathcal{G}_\text{hit}$ represents the set of overlapping Gaussian kernels at position $q$, and $g$ indexes the overlapping kernels.

In regions with multiple overlapping kernels, the minimum value is selected to generate a denser and more continuous multi-view depth map. 
The enhanced depth maps $Y$ are then passed through a shared encoder $\varphi\left(\cdot\right)$ to extract depth features $F_\mathrm{E}^{(m,i)} \in \mathbb{R}^{H_\mathrm{I} \times W_\mathrm{I} \times \mathcal{A}}$ , where $\mathcal{A}$ denotes the dimension of the depth features. 
These features are concatenated with the corresponding 2D image features $F_\mathrm{I}^{(m,i)}$, and the composite features are processed by a depth fusion network $\psi\left(\cdot\right)$ to obtain depth-aware 2D image features $F_{\mathrm{E2I}}^{(m,i)} \in \mathbb{R}^{H_\mathrm{I} \times W_\mathrm{I} \times C_\mathrm{I}}$, thereby enhancing the geometric understanding of the image content. 
The overall process can be formulated as:
\begin{align}
    F_{\mathrm{E2I}}^{(m,i)} &= \psi \left( \mathtt{Concat} \left[ F_\mathrm{E}^{(m,i)}, F_\mathrm{I}^{(m,i)} \right] \right), \\
    F_\mathrm{E}^{(m,i)} &= \varphi \left( Y \right).
\end{align}
Ultimately, the depth-aware 2D image features $F_{\mathrm{E2I}}^{(m,i)}$ are fed into the LSS \cite{lss} module, which transforms them into 3D voxels $V_\mathrm{I} \in \mathbb{R}^{D \times H \times W \times C}$ from image modality, matching the dimensions of the LiDAR voxel features. This transformation enriches geometric-aware 2D image features, yielding a more comprehensive and spatially accurate scene representation.

\subsubsection{Semantic-aware Module} \label{semantic-net}

\begin{figure}[t]
    \centering
    \includegraphics[width=0.6\linewidth]{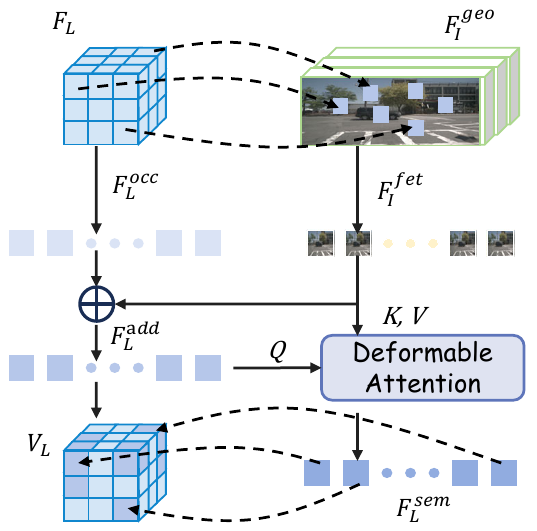}
    \caption{Illustration of the proposed Semantic-Aware module. Semantic information from the camera modality, aligned with geometric features, is transferred to the LiDAR modality to enrich voxel-level representations.}
    \label{fig: semantic-net}
    \vspace{-12pt}
\end{figure}

As shown in Fig.~\ref{fig: semantic-net}, given the voxel features $F_{\mathrm{L}}$, only a limited number of voxels are actually occupied by LiDAR point clouds. 
For each occupied voxel at location $(x, y, z)$, where $x \in \left\{ 0,1,\cdots, W-1 \right\}$, $y \in \left\{ 0,1,\cdots, H-1 \right\}$, and $z \in \left\{ 0,1,\cdots, D-1 \right\}$, the voxel center is projected onto the depth-aware image features of each view, denoted as $F_\mathrm{I}^{\mathrm{geo}}$. 
The corresponding 2D image features $F_\mathrm{I}^{\mathrm{fet}}$ are then extracted at these projected points and combined with the original occupied voxel features $F_\mathrm{L}^{\mathrm{occ}}$ through element-wise addition. 
The resulting aggregated features $F_\mathrm{L}^{\mathrm{add}}$ serve as queries and interact with the multi-scale image features $F_\mathrm{I}^{\mathrm{fet}}$ of the respective image views via a deformable attention module \cite{DeformableDETR, bevformer}. 
This process is performed independently for each image view:
\begin{equation}
    \mathtt{DA}\left(u, r, e \right) = \sum_{a=1}^{N_{\mathrm{head}}} W_a\sum_{b=1}^{N_{\mathrm{key}}} A_{ab} \cdot W_a' e\left(r + \Delta r_{ab}\right).
\end{equation}
Here, $u$ and $e$ denote the queries and keys, respectively, $W_a$ and $W_a'$ are learnable projection matrices, $A_{ab}$ represents the attention weights, $r$ is the 2D reference point onto which the voxel center is projected, and $\Delta r_{ab}$ denotes the positional offset. 
Due to the overlapping fields of multiple views, a voxel may be projected onto several image views. 
To consolidate the updated features across views, max-pooling is applied, resulting in the output feature $F_{\mathrm{L}}^{\mathrm{sem}} = \left\{ F_{\mathrm{L},k}^{\mathrm{sem}} \right\}_{k=1}^{N_\mathrm{v}}$, where $N_\mathrm{v}$ is the number of occupied voxels. Each updated voxel feature $F_{\mathrm{L},k}^{\mathrm{sem}}$ is defined as:
\begin{equation}
    F_{\mathrm{L}, k}^{\mathrm{sem}} = \max_{j \in \mathcal{V}_{hit}} \mathtt{DA} \left(F_\mathrm{L}^{\mathrm{add}}, \mathtt{Proj}\left(q_{\mathrm{3D}}^{k}, j \right), F_\mathrm{I}^{\mathrm{fet}} \right).
\end{equation}
Here, $j$ denotes the index of the image view, $F_\mathrm{L}^{\mathrm{add}}$ is the aggregated query feature for the $k$-th voxel, and $\mathtt{Proj}\left(q_{\mathrm{3D}}^{k}, j \right)$ denotes the projection function mapping the 3D voxel center $q_{\mathrm{3D}}^{k}$ to the $j$-th image view. 
$\mathcal{V}_{hit}$ is the set of views in which the voxel center is visible. 
The refined semantic-aware features $F_\mathrm{L}^{\mathrm{sem}}$ are then used to replace the original features in $F_\mathrm{L}$ at their respective occupied voxel locations $O$, yielding the semantic-aware 3D LiDAR voxel features $V_\mathrm{L}$, defined as:
\begin{align}
    V_{\mathrm{L}} &= \tau \left( F_{\mathrm{L}}^{\mathrm{sem}}, F_{\mathrm{L}}, O \right)  \in \mathbb{R}^{\mathrm{D \times H \times W \times C}}, \\
    O &= \left\{ \mathcal{O}_l \right\}_{l=1}^{N_{\mathrm{occ}}}.
\end{align}
In this context, $O$ represents the set of 3D locations corresponding to $F_\mathrm{L}^{\mathrm{sem}}$, with $N_{\mathrm{occ}}$ occupied voxels in total. Each $\mathcal{O}_l$ denotes the 3D index of the $l$-th occupied voxel, and $\tau(\cdot)$ represents the update operation.

\subsection{Late Stage Voxel Fusion}\label{late-fusion}
The HCCVF and AF modules are designed to perform late-stage voxel fusion to refine 3D semantic occupancy perception.

\subsubsection{High Classification Confidence Voxel Fusion Module} \label{HCC fusion}
The HCCVF module, as illustrated in Fig.~\ref{fig:framework}, first employs a dual-branch prediction head to estimate classification confidence scores for both modalities. 
Top-$K$ ranked voxels from each modality are then selected, denoted as $V_{\mathrm{L}}^{\mathrm{top}}$ and $V_{\mathrm{I}}^{\mathrm{top}}$ for the LiDAR and camera modalities, respectively. 
Beyond filtering high-confidence occupancy predictions, this mechanism also identifies voxels with conflicting classification confidence scores across modalities, which may indicate semantic ambiguity. 
These ambiguous voxels are prioritized for refinement and passed through modality-specific learnable projectors: $\mathrm{f_{L}(\cdot)}$ for $V_{\mathrm{L}}^{\mathrm{top}}$ and $\mathrm{f_{C}(\cdot)}$ for $V_{\mathrm{I}}^{\mathrm{top}}$. 
The projected features are then concatenated along the sequence dimension to form the set $V_{\mathrm{LC}}^{\mathrm{top}}$:
\begin{equation}
    V_{\mathrm{LC}}^{\mathrm{top}} = \mathrm{f_{L}}\left(V_{\mathrm{L}}^{\mathrm{top}}\right) \cup \mathrm{f_{C}}\left(V_{\mathrm{I}}^{\mathrm{top}}\right).
\end{equation}
This concatenated set is subsequently passed through a self-attention module followed by a feed-forward network to obtain the final fused instance features $V_{\mathrm{F}}^{\mathrm{top}}$. Within the self-attention module, learned positional embeddings are incorporated by encoding the XYZ coordinates of the occupancy voxels using a Multi-Layer Perceptron (MLP) and adding them to the instance features.

\subsubsection{Adaptive Fusion Module} \label{adaptive fusion}
As depicted in Fig.~\ref{fig:framework}, the AF module employs an adaptive fusion mechanism based on a lightweight 3D convolutional network to enhance the geometric and semantic features at corresponding voxel locations \cite{wang2023openoccupancy}. This results in the fused 3D occupancy voxel features, denoted as $V_{\mathrm{F}}^{\mathrm{ada}}$. The fusion mechanism is formulated as:
\begin{align}
    V_{\mathrm{F}}^{\mathrm{ada}} &= W_{\mathrm{vis}} \odot V_{\mathrm{I}} + (1 - W_{\mathrm{vis}}) \odot V_{\mathrm{L}}, \\
    W_{\mathrm{vis}} &= \sigma(\mathtt{Conv}(\mathtt{Concat}\left[ V_{\mathrm{L}}, V_{\mathrm{I}} \right])).
\end{align}
Here, $\mathtt{Conv}(\cdot)$ denotes the weight encoding network, which primarily consists of a 3D convolutional network and normalization layers. The weight $W_{\mathrm{vis}}$ is computed via a sigmoid function and is used to balance the contribution of semantic features from the camera and geometric features from LiDAR. Subsequently, the fused instance features $V_{\mathrm{L}}^{\mathrm{top}}$ from the HCCVF module replace part of the features in $V_{\mathrm{F}}^{\mathrm{ada}}$ at the corresponding 3D voxel locations $P$, resulting in the final fused 3D occupancy voxel features $V_{\mathrm{F}}$:
\begin{align}
    V_{\mathrm{F}} &= \chi \left( V_{\mathrm{F}}^{\mathrm{top}}, V_{\mathrm{F}}^{\mathrm{ada}}, P \right), \\
    P &= \left\{ \mathcal{P}_h \right\}_{h=1}^{N_{\mathrm{LC}}}.
\end{align}
In this formulation, $P$ denotes the set of 3D locations corresponding to $V_{\mathrm{F}}^{\mathrm{top}}$, and $\chi(\cdot)$ denotes the replacement operation. Finally, a 3D convolutional decoder is applied to the fused features $V_{\mathrm{F}}$ to produce the final occupancy prediction results.

\subsection{Loss Function} \label{objective-function}
To train the proposed MS-Occ, the cross-entropy loss $\mathcal{L}_\text{ce}$ and the Lovász-Softmax loss $\mathcal{L}_\text{ls}$ \cite{lovasz-softmax} are utilized. Following \cite{monoscene}, affinity losses $\mathcal{L}_\text{scal}^\text{geo}$ and $\mathcal{L}_\text{scal}^\text{sem}$ are also employed. In addition, explicit depth supervision $\mathcal{L}_\text{d}$ \cite{bevdepth} is introduced to guide the depth-aware view transformation module. An auxiliary mask supervision is applied using focal loss, denoted as $\mathcal{L}_\text{mask}$. The total loss is formulated as:
\begin{align}
    \mathcal{L}_\mathrm{total} &= \mathcal{L}_\mathrm{ce} + \mathcal{L}_\mathrm{ls} + \mathcal{L}_\mathrm{scal}^\mathrm{geo} + \mathcal{L}_\mathrm{scal}^\mathrm{sem} + \mathcal{L}_\mathrm{d} + \mathcal{L}_\mathrm{mask}, \\
    \mathcal{L}_\mathrm{mask} &= \mathcal{L}_\mathrm{Focal}(\hat{V_{\mathrm{L}}}, \mathcal{V_{\mathrm{L}}}) + \mathcal{L}_\mathrm{Focal}(\hat{V_{\mathrm{I}}}, \mathcal{V_{\mathrm{I}}}),
\end{align}
where $\hat{V_{\mathrm{L}}}$ and $\hat{V_{\mathrm{I}}}$ are the classification predictions, while $\mathcal{V_{\mathrm{L}}}$ and $\mathcal{V_{\mathrm{I}}}$ represent their corresponding ground truth voxel labels.

\FloatBarrier 

\section{Experiments}
\subsection{Dataset and Metrics}
Our method is evaluated on two widely accepted, large-scale benchmarks for 3D semantic occupancy prediction. The primary dataset is nuScenes-OpenOccupancy~\cite{wang2023openoccupancy, nuscenes2019}, which comprises 1,000 driving scenarios, annotated with 3D occupancy ground truth provided by~\cite{wang2023openoccupancy}, using a dense voxel grid of size $(512, 512, 40)$. 
A total of 17 classes are defined, including 16 semantic categories and one class representing free space.
To assess generalization capability, we further evaluate on the SemanticKITTI dataset [38], comprising 22 sequences with a voxel grid of size $(256, 256, 32)$, annotated with 21 total classes (19 semantic, 1 free, and 1 unknown).
SemanticKITTI [38] presents a more challenging benchmark featuring denser, complex, and unstructured natural environments with significant terrain and vegetation variations.
Both benchmarks utilize a voxel resolution of (0.2\,m, 0.2\,m, 0.2\,m).
Following \cite{wang2023openoccupancy}, scene completion intersection over union (IoU) is used to evaluate geometric accuracy, while mean IoU (mIoU) is adopted for semantic accuracy.

\vspace{-6pt}
\subsection{Implementation Details}

To ensure a fair comparison, our implementation and key hyperparameter choices closely follow the experimental settings established by prior benchmarks \cite{wang2023openoccupancy, Co-Occ}.
For the camera branch, input images are resized to $1600 \times 900$ and processed using a pretrained ResNet-50 \cite{Resnet} backbone with an FPN \cite{lin2017fpn}. 
For the LiDAR branch, 10 LiDAR sweeps are voxelized and encoded using a voxel-based encoder \cite{Zhou_Tuzel_VoxelNet, SECOND}. 
To balance predictive accuracy and computational efficiency, the Gaussian kernel size $\sigma$ is set as 6, a choice validated through ablation studies (Table~\ref{tab:ablation_total}).
The enhanced depth heatmaps from the Gaussian-Geo module are processed using $1 \times 1$ convolutional layers, and the fused depth features are subsequently encoded by a ResNet-50 \cite{Resnet} backbone. 
In the late fusion stage, a 3D decoder with an upsampling stride of 2 is applied.
The model is implemented using MMDetection3D \cite{contributors2020mmdetection3d} and trained for 24 epochs with a batch size of 8 on 8 NVIDIA A100 GPUs. We use the AdamW optimizer \cite{adamw} with a weight decay of 0.01 and an initial learning rate of $3 \times 10^{-4}$.

\definecolor{nbarrier}{RGB}{112, 128, 144}
\definecolor{nbicycle}{RGB}{220, 20, 60}
\definecolor{nbus}{RGB}{255, 127, 80}
\definecolor{ncar}{RGB}{255, 158, 0}
\definecolor{nconstruction}{RGB}{233, 150, 70}
\definecolor{nmotorcycle}{RGB}{255, 61, 99}
\definecolor{npedestrian}{RGB}{0, 0, 230}
\definecolor{ntrafficcone}{RGB}{47, 79, 79}
\definecolor{ntrailer}{RGB}{255, 140, 0}
\definecolor{ntruck}{RGB}{255, 99, 71}
\definecolor{nmanmade}{RGB}{128, 128, 128}
\definecolor{ndriveable_surface}{RGB}{0, 207, 191}
\definecolor{nother flat}{RGB}{175, 0, 75}
\definecolor{nsidewalk}{RGB}{75, 0, 75}
\definecolor{nterrain}{RGB}{112, 180, 60}
\definecolor{nvegetation}{RGB}{0, 175, 0}

\setlength{\tabcolsep}{1.4pt}
\renewcommand{\arraystretch}{1}

\begin{table*}[t]
\centering
\caption{Comparison with state-of-the-arts on nuScenes-OpenOccupancy validation set \cite{nuscenes2019, wang2023openoccupancy}. ``C" and ``L" denote camera and LiDAR.}
\label{tab:occ_performance}
\resizebox{0.95\textwidth}{!}{ 
\begin{tabular}{c|c|cc|ccc|ccc|ccccc|ccccc}
\hline
\noalign{\smallskip}
Methods & Modality & IoU & mIoU 
& \rotatebox{90}{\textcolor{npedestrian}{$\blacksquare$}pedestrian} 
& \rotatebox{90}{\textcolor{nbicycle}{$\blacksquare$}bicycle} 
& \rotatebox{90}{\textcolor{nmotorcycle}{$\blacksquare$}motorcycle} 
& \rotatebox{90}{\textcolor{ntrafficcone}{$\blacksquare$}trafficcone}
& \rotatebox{90}{\textcolor{nbarrier}{$\blacksquare$}barrier} 
& \rotatebox{90}{\textcolor{nmanmade}{$\blacksquare$}manmade}
& \rotatebox{90}{\textcolor{ncar}{$\blacksquare$}car}
& \rotatebox{90}{\textcolor{nbus}{$\blacksquare$}bus}
& \rotatebox{90}{\textcolor{ntruck}{$\blacksquare$}truck}
& \rotatebox{90}{\textcolor{ntrailer}{$\blacksquare$}trailer}
& \rotatebox{90}{\textcolor{nconstruction}{$\blacksquare$}const. veh.}
& \rotatebox{90}{\textcolor{ndriveable_surface}{$\blacksquare$}drive. surf.}
& \rotatebox{90}{\textcolor{nsidewalk}{$\blacksquare$}sidewalk}
& \rotatebox{90}{\textcolor{nterrain}{$\blacksquare$}terrain}
& \rotatebox{90}{\textcolor{nother flat}{$\blacksquare$}other flat}
& \rotatebox{90}{\textcolor{nvegetation}{$\blacksquare$}vegetation}
\\
\noalign{\smallskip}
\hline
\noalign{\smallskip}
TPVFormer (CVPR 2023) \cite{TPVFormer} & C & 15.3 & 7.8 & 5.9 & 4.1 & 4.3 & 5.3 & 9.3 & 9.2 & 10.1 & 11.3 & 6.5 & 6.8 & 5.2 & 13.6 & 8.3 & 8.0 & 9.0 & 8.2 \\
SurroundOcc (ICCV 2023) \cite{surroundocc} & C & 31.4 & 20.3 & 14.0 & 11.6 & 15.1 & 12.0 & 20.5 & 14.8 & \textbf{30.8} & 28.1 & 22.2 & 14.3 & 10.7 & 37.2 & 24.4 & 22.7 & 23.7 & 21.8 \\
OccFormer (ICCV 2023) \cite{zhang2023occformer} & C & 29.9 & 20.1 & 14.4 & 11.3 & 15.7 & 11.2 & 21.1 & 15.2 & 30.3 & \textbf{28.2} & 22.6 & 14.0 & 10.6 & 37.3 & 24.9 & 23.5 & 22.4 & 21.1 \\
LMSCNet (3DV 2020) \cite{lmscnet} & L & 27.3 & 11.5 & 6.2 & 4.2 & 4.7 & 6.3 & 12.4 & 13.9 & 12.1 & 12.8 & 7.2 & 8.8 & 6.2 & 24.2 & 16.6 & 14.1 & 12.3 & 22.2 \\
L-CONet (ICCV 2023) \cite{wang2023openoccupancy} & L & 30.9 & 15.8 & 9.6 & 5.2 & 5.4 & 5.6 & 17.5 & 19.2 & 18.1 & 13.3 & 13.6 & 13.2 & 7.8 & 34.9 & 22.4 & 21.7 & 21.5 & 23.5 \\
M-CONet (ICCV 2023) \cite{wang2023openoccupancy} & C\&L & 29.5 & 20.1 & 18.0 & 13.3 & 15.9 & 13.3 & 23.3 & 19.6 & 24.3 & 21.2 & 20.7 & 15.3 & 15.3 & 33.2 & 22.5 & 21.5 & 21.0 & 23.2 \\
Co-Occ (IEEE RA-L 2024) \cite{Co-Occ} & C\&L & 30.6 & 21.9 & 20.7 & 16.8 & 20.9 & 14.5 & 26.5 & 20.5 & 27.0 & 22.3 & 21.6 & 16.4 & 10.1 & 36.9 & \textbf{25.5} & 23.7 & 23.5 & 23.5 \\
OccGen (ECCV 2024) \cite{OccGen} & C\&L & 30.3 & 22.0 & 21.6 & 16.4 & 20.1 & 14.6 & 24.9 & 20.5 & 26.1 & 22.5 & 21.9 & \textbf{17.4} & 14.0 & 35.8 & 24.7 & 24.0 & \textbf{24.5} & 23.5 \\
OccLoff (WACV 2025) \cite{occloff} & C\&L & 31.4 & 22.9 & 24.7 & 17.2 & 22.6 & 16.4 & 26.7 & 21.4 & 26.9 & 22.6 & 22.0 & 16.3 & 16.4 & \textbf{37.5} & 25.3 & 23.9 & 22.3 & 24.2 \\
\noalign{\smallskip}
\hline
\noalign{\smallskip}
\textbf{MS-Occ(ours)} & C\&L & \textbf{32.1} & \textbf{25.3} & \textbf{29.0} & \textbf{23.8} & \textbf{27.5} & \textbf{21.1} & \textbf{29.4} & \textbf{27.6} & 28.7 & 24.2 & \textbf{25.3} & 16.9 & \textbf{17.5} & 33.4 & 24.4 & \textbf{24.6} & 23.5 & \textbf{28.4} \\
\noalign{\smallskip}
\hline
\end{tabular}}
\end{table*}
\vspace{-4pt}
\definecolor{scar}{rgb}{0.39215686, 0.58823529, 0.96078431}
\definecolor{sbicycle}{rgb}{0.39215686, 0.90196078, 0.96078431}
\definecolor{smotorcycle}{rgb}{0.11764706, 0.23529412, 0.58823529}
\definecolor{struck}{rgb}{0.31372549, 0.11764706, 0.70588235}
\definecolor{sother-vehicle}{rgb}{0.39215686, 0.31372549, 0.98039216}
\definecolor{sperson}{rgb}{1.        , 0.11764706, 0.11764706}
\definecolor{sbicyclist}{rgb}{1.        , 0.15686275, 0.78431373}
\definecolor{smotorcyclist}{rgb}{0.58823529, 0.11764706, 0.35294118}
\definecolor{sroad}{rgb}{1.        , 0.        , 1.        }
\definecolor{sparking}{rgb}{1.        , 0.58823529, 1.        }
\definecolor{ssidewalk}{rgb}{0.29411765, 0.        , 0.29411765}
\definecolor{sother-ground}{rgb}{0.68627451, 0.        , 0.29411765}
\definecolor{sbuilding}{rgb}{1.        , 0.78431373, 0.        }
\definecolor{sfence}{rgb}{1.        , 0.47058824, 0.19607843}
\definecolor{svegetation}{rgb}{0.        , 0.68627451, 0.        }
\definecolor{strunk}{rgb}{0.52941176, 0.23529412, 0.        }
\definecolor{sterrain}{rgb}{0.58823529, 0.94117647, 0.31372549}
\definecolor{spole}{rgb}{1.        , 0.94117647, 0.58823529}
\definecolor{straffic-sign}{rgb}{1.        , 0.        , 0.        }


\setlength{\tabcolsep}{1.4pt}  
\renewcommand{\arraystretch}{1}  

\begin{table*}[t]
\centering
\caption{Comparison with state-of-the-arts on the SemanticKITTI validation set \cite{semantickitti}. Classes are grouped to highlight performance on VRUs, Small Objects, and Complex Structures.
``C" and ``L" denote camera and LiDAR.
}
\resizebox{0.95\textwidth}{!}{
\begin{tabular}{c|c|c|c c c c c c|c c c|c c c|c c c c c c c}
\hline
\noalign{\smallskip}
    Methods & Modality & mIoU  
    & \rotatebox{90}{\textcolor{sperson}{$\blacksquare$}  person}
    & \rotatebox{90}{\textcolor{sbicyclist}{$\blacksquare$}  bicyclist}
    & \rotatebox{90}{\textcolor{smotorcyclist}{$\blacksquare$}  motorcyclist.}
    & \rotatebox{90}{\textcolor{sbicycle}{$\blacksquare$}  bicycle}
    & \rotatebox{90}{\textcolor{smotorcycle}{$\blacksquare$} motorcycle}
    & \rotatebox{90}{\textcolor{straffic-sign}{$\blacksquare$} traf.-sign}
    & \rotatebox{90}{\textcolor{scar}{$\blacksquare$}  car}
    & \rotatebox{90}{\textcolor{struck}{$\blacksquare$}  truck}
    & \rotatebox{90}{\textcolor{sother-vehicle}{$\blacksquare$}  other-veh.}
    & \rotatebox{90}{\textcolor{svegetation}{$\blacksquare$} vegetation}
    & \rotatebox{90}{\textcolor{strunk}{$\blacksquare$}  trunk}
    & \rotatebox{90}{\textcolor{sfence}{$\blacksquare$} fence}
    & \rotatebox{90}{\textcolor{sroad}{$\blacksquare$} road}
    & \rotatebox{90}{\textcolor{ssidewalk}{$\blacksquare$} sidewalk}
    & \rotatebox{90}{\textcolor{sparking}{$\blacksquare$} parking}
    & \rotatebox{90}{\textcolor{sother-ground}{$\blacksquare$} other-grnd}
    & \rotatebox{90}{\textcolor{sbuilding}{$\blacksquare$}  building}
    & \rotatebox{90}{\textcolor{sterrain}{$\blacksquare$} terrain}
    & \rotatebox{90}{\textcolor{spole}{$\blacksquare$} pole}
    \\
\noalign{\smallskip}
\hline
\noalign{\smallskip}

JS3C-Net (AAAI 2021)~\cite{JS3C-Net}&C  &10.31& 0.67 &0.27 & 0.00 & 0.00 & 0.00& 1.45 &24.65 & 4.41 & 6.15  &18.11&4.33 & 3.94 & 50.49& 23.74 & 11.94 & 0.07& 15.03 & 26.86 & 3.77    \\
MonoScene (CVPR 2022)~\cite{monoscene}&C  & 11.08 & 1.86 & 1.20 & 0.00 & 0.61 & 0.45 & 2.25 & 23.26 & 6.98 & 1.48& 17.89 & 2.81 & 5.84 & 56.52 & 26.72 &  14.27& 0.46 & 14.09 & 29.64     & 4.14  \\
VoxFormer (CVPR 2023)~\cite{li2023voxformer}&C  & 12.35 & 1.78 & 3.32 & 0.00 & 0.59 & 0.51 & 4.18 & 25.79 & 5.63 & 3.77 & 24.39 & 5.08 & 7.64& 54.76 & 26.35 & 15.50 & \textbf{0.70} &17.65 & 29.96 & 7.11 \\ 
LMSCNet (3DV 2020)~\cite{lmscnet} &L &6.70& 0.00& 0.00& 0.00& 0.0& 0.0& 0.00& 18.33 &0.00& 0.0& 13.66& 0.02 & 1.21&40.68& 18.22 &4.38& 0.00& 10.31 &20.54& 0.00\\

M-CONet (CVPR 2023)~\cite{wang2023openoccupancy}&C\&L  & 17.25 & 2.07 & 2.38 & 0.00 &0.66 &2.2 & 14.09 &33.25 & 19.46 & 5.99 & 32.56 &16.01 & 12.25 & 58.00 & 29.44 &19.33  & 0.02 & 27.85 &37.31 &14.88  \\
OccFusion (IEEE T-IV 2024)~\cite{occfusion}&C\&L& 21.64& 2.33& 3.07& 0.00& 2.00 &3.39& 14.36&45.72 &19.41 & 7.92& 41.21 &19.18 & 15.74&65.73& 36.46& \textbf{21.35} &0.00 &39.32 &\textbf{46.28} & \textbf{27.60}\\

OccLoff (WACV 2025)~\cite{occloff} &C\&L & 22.62 & 3.88 & 4.35& 0.00 &2.08&3.91& 16.47 & 46.44 & 20.38&8.72&41.20&20.06 & 15.86 &\textbf{66.25}  &\textbf{43.51} & 21.07 & 0.57 & 41.23 &46.21& \textbf{27.60}   \\
\noalign{\smallskip}
\hline
\noalign{\smallskip}

\textbf{MS-Occ} (Ours)&C\&L & \textbf{24.08} & \textbf{5.32} & \textbf{6.59}& 0.00 &\textbf{3.96 }&\textbf{4.65 }& \textbf{18.61} & \textbf{47.78} & \textbf{22.82 }&\textbf{10.86}&\textbf{46.44}&\textbf{24.7} & \textbf{18.6} &64.89  &42.65 & 20.81 & 0.51 &\textbf{44.67} &46.15 & 27.44   \\

\noalign{\smallskip}
\hline
\end{tabular}}
\label{tab:occ_performance_semanticKITTI}
\end{table*}
\setlength{\tabcolsep}{1.4pt}
\vspace{-4pt}





\begin{table}[ht] 
\centering
\caption{Comparison of accuracy and efficiency with state-of-the-art methods on the nuScenes validation set. All Frames Per Second (FPS) benchmarks were conducted on a single NVIDIA A100 GPU for a fair comparison.}
    \label{tab:efficiency_comparison} 
    
    \resizebox{\columnwidth}{!}{%
        \begin{tabular}{l | c c c | c c}
    \toprule
    \makebox[2.5cm]{\textbf{Method}} & 
    \makebox[1.5cm]{\textbf{Params (M)}} & 
    \makebox[1.0cm]{\textbf{GFLOPs}} & 
    \makebox[1.0cm]{\textbf{FPS}} & 
    \makebox[1.0cm]{\textbf{IoU}} & 
    \makebox[1.0cm]{\textbf{mIoU}} \\
    \midrule
    M-OpenOccupancy & 123.37 & 3045.11 & 2.3 & 29.1 & 15.1 \\
    M-CONet & 123.45 & 3089.47 & 1.6 & 29.5 & 20.1 \\
    \midrule
    \textbf{MS-Occ (Ours)} & \textbf{100.8} & \textbf{3617.9} & \textbf{1.3} & \textbf{32.1} & \textbf{25.3} \\
    \bottomrule
    \end{tabular}%
    }
\end{table}

\vspace{-4pt}

\begin{figure*}[!t]
    \centering
    \includegraphics[width=0.85\linewidth]{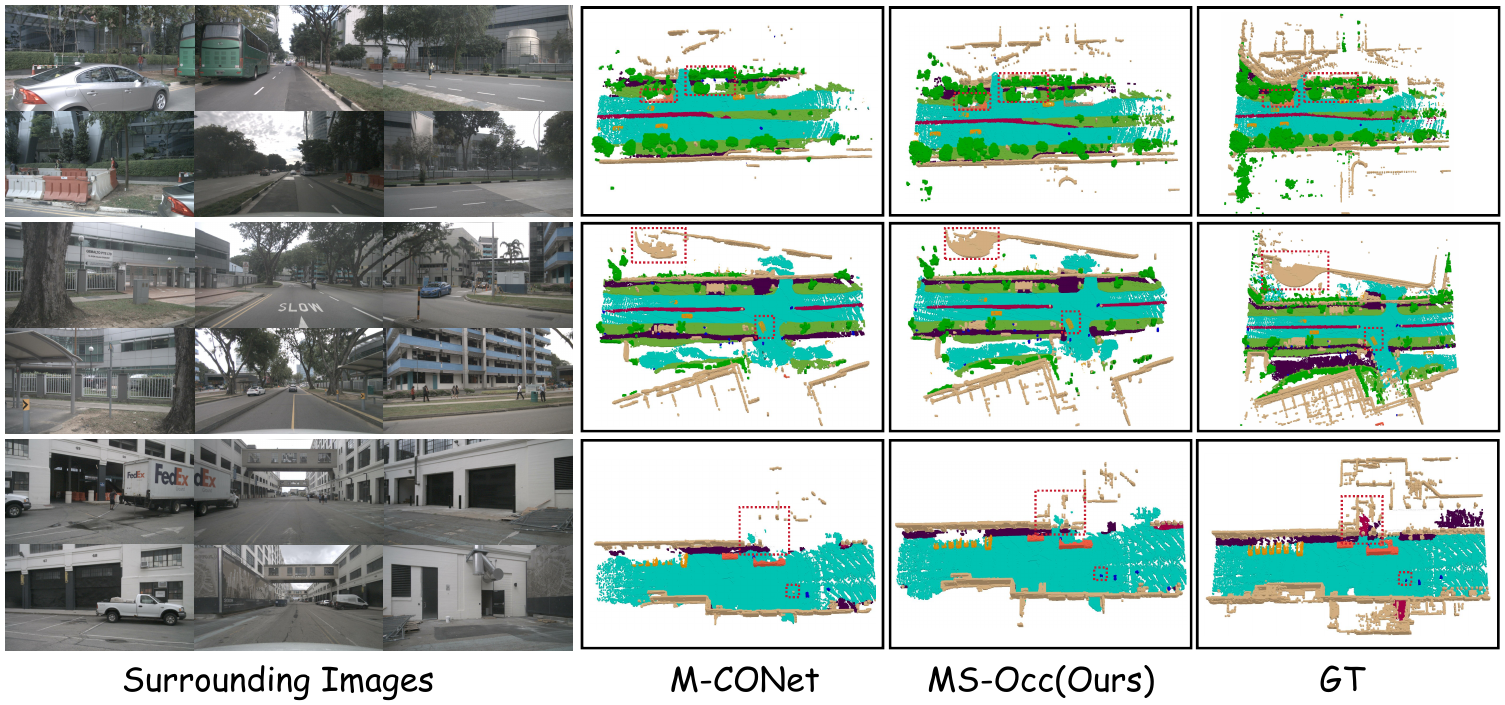}
    \caption{Visualization results on the nuScenes-OpenOccupancy dataset \cite{nuscenes2019, wang2023openoccupancy}. The leftmost column shows the surround-view images. The next three columns present 3D semantic occupancy predictions from M-CONet, MS-Occ (ours), and the ground truth. Please zoom in for finer details.}
    \label{fig:Visualization}
    \vspace{-12pt}
\end{figure*}

\subsection{Main Results}
\subsubsection{Results on NuScenes Dataset}
As quantitatively shown in Table~\ref{tab:occ_performance}, the proposed MS-Occ framework achieves superior overall performance with an IoU of 32.1 and an mIoU of 25.3, demonstrating particularly significant gains in vulnerable road users (VRUs) categories (pedestrians, bicycles, motorcycles) and small object detection (traffic cones, man-made structures). Compared to the baseline M-CONet~\cite{wang2023openoccupancy}, these critical classes exhibit substantial IoU improvements of 11.0\%, 10.5\%, 11.6\%, 7.8\%, and 8.0\%, respectively. These improvements are especially important for autonomous driving safety due to the high kinematic uncertainty of VRUs and the low reflectivity of traffic cones.
By fusing rich visual context, MS-Occ effectively mitigates the point sparsity limitations of LiDAR-only methods, elevating performance on challenging classes, such as pedestrian detection from 9.6\% (L-CONet) to a superior 29.0\% mIoU.

In comparison with state-of-the-art LiDAR-camera fusion methods, MS-Occ consistently outperforms across different fusion stages. Specifically, it achieves a 2.4\% mIoU improvement over OccLoff~\cite{occloff}, a representative middle-stage fusion method. Moreover, compared to Co-Occ, a leading late-stage fusion approach, MS-Occ yields even greater improvements of 3.4\% in mIoU and 1.5\% in IoU. Fig.~\ref{fig:Visualization} illustrates that MS-Occ produces 3D occupancy predictions with fewer false positives and improved boundary delineation. The color scheme used in Fig.~\ref{fig:Visualization} corresponds to the class columns in Table~\ref{tab:occ_performance}.

\subsubsection{Results on SemanticKITTI}
To further validate the generalization capability, we evaluate our model on the SemanticKITTI dataset \cite{semantickitti}. 
As shown in Table~\ref{tab:occ_performance_semanticKITTI}, MS-Occ establishes a new state-of-the-art mIoU of 24.08, substantially outperforming prior methods such as M-CONet (17.25) and OccLoff (22.62). 
This result demonstrates that the proposed multi-stage fusion architecture is broadly effective and does not overfit a single dataset.
Notably, Table~\ref{tab:occ_performance_semanticKITTI} highlights the strong performance of MS-Occ on challenging small objects (e.g., bicyclist, person, traffic-sign) and geometrically complex structures (e.g., vegetation, trunk, fence).

\subsubsection{Efficiency Analysis}
Table~\ref{tab:efficiency_comparison} summarizes the efficiency analytsis.
MS-Occ attains state-of-the-art IoU and mIoU while maintaining competitive inference speed and requiring 18.3\% fewer parameters. 
This reflects a highly favorable trade-off, achieving critical accuracy gains for driving safety within a more compact and parameter-efficient architecture.

\subsection{Ablation Study}
%
%
To validate the effectiveness of the proposed components, we perform a series of ablation studies, all conducted on the nuScenes-OpenOccupancy validation set.

\begin{table}[ht] %
\centering
\caption{Ablation studies on fusion stages, key components, and design choices. "w/o" denotes the removal of a module/stage.}
\label{tab:ablation_total} %

\resizebox{\columnwidth}{!}{%
    \begin{tabular}{c|cc|cccc|c|cc}
    \toprule
    \textbf{Exp.} & \textbf{Mid. Stage} & \textbf{Late Stage} & \textbf{G-Geo} & \textbf{Sem-Aware} & \textbf{HCCVF} & \textbf{AF} & \textbf{Kernel} & \textbf{IoU} & \textbf{mIoU} \\
    \midrule
    A & \checkmark & w/o Late & \checkmark & \checkmark & - & - & 6 & 31.4 & 23.9 \\
    B & w/o Middle & \checkmark & - & - & Self-Attention & \checkmark & - & 30.2 & 21.1 \\
    \midrule
    C & \checkmark & \checkmark & w/o GG & \checkmark & Self-Attention & \checkmark & - & 31.7 & 24.4 \\
    D & \checkmark & \checkmark & \checkmark & w/o SA & Self-Attention & \checkmark & 6 & 30.6 & 22.1 \\
    E & \checkmark & \checkmark & \checkmark & \checkmark & w/o HCCVF & \checkmark & 6 & 31.6 & 24.3 \\
    F & \checkmark & \checkmark & \checkmark & \checkmark & Self-Attention & w/o AF & 6 & 31.9 & 24.9 \\
    G & \checkmark & \checkmark & \checkmark & \checkmark & Summation & \checkmark & 6 & 31.6 & 24.5 \\
    H & \checkmark & \checkmark & \checkmark & \checkmark & Concatenation & \checkmark & 6 & 31.7 & 24.7 \\
    \midrule
    I & \checkmark & \checkmark & \checkmark & \checkmark & Self-Attention & \checkmark & 3 & 31.8 & 24.7 \\
    J & \checkmark & \checkmark & \checkmark & \checkmark & Self-Attention & \checkmark & 10 & 32.2 & 25.5 \\
    \midrule
    K & \checkmark & \checkmark & \checkmark & \checkmark & Self-Attention & \checkmark & \textbf{6} & \textbf{32.1} & \textbf{25.3} \\
    \bottomrule
    \end{tabular}%
}
\end{table}
\vspace{-6pt}
\textbf{Effect of Fusion Stages.}
Ablation studies are conducted to evaluate the impact of different fusion stages on the overall model performance, as presented in Table~\ref{tab:ablation_total} (Exp. A-B). The middle-stage feature fusion effectively compensates for the semantic deficiencies of the LiDAR modality and the geometric limitations of the camera modality. Removing this stage results in a significant performance drop of 4.2\% in mIoU. The voxel fusion in the late stage primarily refines feature integration in 3D space. Eliminating this stage leads to a 1.4\% decrease in mIoU.

\textbf{Effect of Main Components.} 
Table~\ref{tab:ablation_total} (Exp. C-F) presents the impact of key components on model performance. The ablation results reveal the following: (1) Removing the Gaussian-Geo module results in a 0.9\% decrease in mIoU; (2) Eliminating the Semantic-Aware module causes a substantial 3.2\% drop in mIoU; (3) Discarding the AF module leads to a 0.4\% decline in mIoU; and (4) Removing the HCCVF module results in a 1.0\% decrease in mIoU. These demonstrate that each component contributes meaningfully to MS-Occ framework.

\textbf{Ablation of HCCVF.}
Table~\ref{tab:ablation_total} (Exp. G-H) presents the ablation study on fusion strategies in the late fusion stage. To evaluate its effectiveness, HCCVF is compared with two alternative fusion methods adapted from \cite{MultiModalObjectDetection}: (1) Summation-based fusion and (2) Concatenation-based fusion. The experimental results demonstrate that the HCCVF module achieves superior performance compared to these conventional fusion strategies. Specifically, HCCVF outperforms the summation-based method by 0.8\% in mIoU and surpasses the concatenation-based method by 0.6\% in mIoU.

\textbf{Ablation of Gaussian Kernel.} 
Table~\ref{tab:ablation_total} (Exp. I-J) presents an ablation study on the Gaussian kernel length, $\sigma$, in the Gaussian-Geo module. A significant improvement is observed as $\sigma$ increases from 3 to 6, while further increasing $\sigma$ to 10 yields only marginal gains. This suggests that the influence of kernel size reaches a saturation point. Given that larger kernel sizes incur higher memory consumption, $\sigma = 6$ is selected to achieve the best balance between performance and computational efficiency.

\section{Conclusions}
In this letter, we introduce MS-Occ, a multi-stage LiDAR-camera fusion framework for 3D semantic occupancy. MS-Occ couples middle-stage feature fusion with late-stage voxel fusion to achieve strong geometric and semantic alignment. Gaussian-Geo injects geometric priors into image features, Semantic-Aware enriches LiDAR voxels, and AF and HCCVF refine voxel representations to resolve semantic ambiguities.
Experiments on nuScenes and SemanticKITTI confirm the effectiveness of MS-Occ, with notable gains in detecting VRUs and other small, safety-critical objects.
Nevertheless, a limitation is that performance gains are less pronounced for large objects. This is likely because strong visual cues already make camera-only methods highly competitive in these cases, whereas our fusion mechanism is mainly designed to enhance fine-grained details and resolve geometric ambiguities in smaller or complex objects.
Future work could explore multi-scale fusion to better capture structural information from large objects. 
Furthermore, the current framework assumes accurate LiDAR-camera extrinsic calibration. 
While our model shows some robustness, improving its resilience to real-world sensor misalignments is an important future direction.

\bibliographystyle{IEEEtran}
\footnotesize
\bibliography{IEEEexample}

\end{document}